%% file: main.tex
\documentclass[11pt]{article}

\usepackage[final]{acl}
\usepackage{booktabs} 
\usepackage{multirow} 
\usepackage{graphicx}
\usepackage{times}
\usepackage{latexsym}
\usepackage{amsmath}
\usepackage{amssymb}
\usepackage{mathtools}
\usepackage{tabularx}
\usepackage[T1]{fontenc}
\usepackage{hyperref}

\usepackage[utf8]{inputenc}

\usepackage{microtype}

\usepackage{inconsolata}

\usepackage{graphicx}
\usepackage{kotex}
\title{Enhancing Low-Resource Language Reasoning via High-Resource Language Feature Transfer}

\author{
    Minju Song$^{1}$ \hspace*{0.1cm}
    Hyeon Hwang$^{1}$ \hspace*{0.1cm}
    Junhyun Lee$^{2,3}$\thanks{Corresponding author} \hspace*{0.1cm}
    Jaewoo Kang$^{1,4}$\footnotemark[1] \\[0.5pt]
    Korea University$^{1}$ \hspace*{0.1cm}
    Hankuk University of Foreign Studies$^{2}$ \\
    Noah's Farm$^{3}$ \hspace*{0.1cm}
    AIGEN Sciences$^{4}$ \\
     \{minjusong, hyeon-hwang, kangj\}@korea.ac.kr \hspace*{0.1cm}
    junhyun.lee@hufs.ac.kr \\
}

\begin{document}
\maketitle

\input{sections/0_abstract}
\input{sections/1_introduction}
\input{sections/2_related}
\input{sections/3_method}
\input{sections/4_experiments}

\input{sections/5_discussion}

\input{sections/6_related_work}
\input{sections/7_conclusion}

\section*{Limitations}

\paragraph{Models and benchmarks.}
We evaluate two instruction-tuned models (Gemma-2-9B-it, Qwen2.5-7B-Instruct) on three benchmarks (MATH500, MGSM, and the psychology subset of MMLU-ProX) covering four low-resource target languages. Generalization to larger or differently architected models, to other reasoning domains (e.g., commonsense, code), and to truly extremely low-resource languages remains to be verified.

\paragraph{Reliance on pretrained SAEs.}
Our feature library is constructed in the basis of an externally trained SAE applied at a single layer (layer 20). Both the resolution of the identified features and the effectiveness of steering are therefore bounded by the quality of the underlying SAE; polysemantic or under-trained latents may dilute the signal isolated by our contrastive procedure.

\paragraph{Transfer across tasks.}
Our experiments primarily evaluate feature transfer across languages within the same task. Cross-task transfer may depend on differences in task distributions and the extent to which reasoning mechanisms are shared across datasets, which we do not explicitly model in this work. Establishing when task-specific features transfer across benchmarks therefore remains an important direction for future work.

% This document does not cover the content requirements for ACL or any
% other specific venue.  Check the author instructions for
% information on
% maximum page lengths, the required ``Limitations'' section,
% and so on.

\section*{Acknowledgments}
This work was supported in part by the National Research Foundation of Korea [NRF-2023R1A2C3004176], the Ministry of Health \& Welfare, Republic of Korea [HR20C002103], the Ministry of Science and ICT (MSIT) [RS-2023-00262002], the Institute of Information \& Communications Technology Planning \& Evaluation(IITP)-ICT Creative Consilience Program grant funded by the Korea government(MSIT)(IITP-2026-RS-2020-II201819), and the National Research Foundation of Korea(NRF) grant funded by the Korea governmant(MSIT and MOE) (No. RS-2025-16652968). The work of Junhyun Lee was supported by the Hankuk University of Foreign Studies Research Fund (2026).    

% This document has been adapted
% by Steven Bethard, Ryan Cotterell and Rui Yan
% from the instructions for earlier ACL and NAACL proceedings, including those for
% ACL 2019 by Douwe Kiela and Ivan Vuli\'{c},
% NAACL 2019 by Stephanie Lukin and Alla Roskovskaya,
% ACL 2018 by Shay Cohen, Kevin Gimpel, and Wei Lu,
% NAACL 2018 by Margaret Mitchell and Stephanie Lukin,
% Bib\TeX{} suggestions for (NA)ACL 2017/2018 from Jason Eisner,
% ACL 2017 by Dan Gildea and Min-Yen Kan,
% NAACL 2017 by Margaret Mitchell,
% ACL 2012 by Maggie Li and Michael White,
% ACL 2010 by Jing-Shin Chang and Philipp Koehn,
% ACL 2008 by Johanna D. Moore, Simone Teufel, James Allan, and Sadaoki Furui,
% ACL 2005 by Hwee Tou Ng and Kemal Oflazer,
% ACL 2002 by Eugene Charniak and Dekang Lin,
% and earlier ACL and EACL formats written by several people, including
% John Chen, Henry S. Thompson and Donald Walker.
% Additional elements were taken from the formatting instructions of the \emph{International Joint Conference on Artificial Intelligence} and the \emph{Conference on Computer Vision and Pattern Recognition}.


\bibliography{custom}

\input{sections/X_appendix}

\end{document}

%% file: sections/0_abstract.tex
% \begin{abstract}
% Large language models exhibit substantial performance variation across languages, even when solving semantically equivalent tasks. 
% Existing analyses often treat this phenomenon as an observational disparity caused by differences in pretraining data, tokenization, or benchmark coverage.
% In particular, high-resource languages may more reliably activate latent computations needed for mathematical reasoning, while lower-resource languages may fail to trigger those mechanisms despite expressing the same task. 
% To test this hypothesis, we introduce a mechanistic intervention framework for identifying and transferring task-relevant sparse latent features across languages. 
% Using sparse autoencoders over residual-stream activations, we isolate features enriched in successful high-resource task-specific reasoning while filtering out source-language and generic-generation features. 
% We then construct steering directions from these features and inject them during lower-resource language inference. 
% This enables direct causal validation: suppressing the discovered features should impair source-language reasoning, while activating them should improve target-language reasoning beyond random and non-task controls. 
% Our framework reframes reasoning gaps across languages as failures of mechanism elicitation rather than merely capability absence, and offers a causally testable route to cross-lingual reasoning transfer without translation, fine-tuning, or changing the user-facing language.
% \end{abstract}

\begin{abstract}
Large language models exhibit substantial performance variation across languages, even when solving semantically equivalent tasks. 
Existing analyses often treat this phenomenon as an observational disparity caused by differences in pretraining data, tokenization, or benchmark coverage.
We study a complementary hypothesis: high-resource languages (HRLs) may more reliably elicit latent computations useful for task-specific (i.e. mathematical) reasoning, while lower-resource languages (LRLs) may under-activate those computations despite expressing the same task.
To test this hypothesis, we introduce a mechanistic intervention framework for identifying and transferring task-relevant sparse latent features across languages.
Using sparse autoencoders over residual-stream activations, we isolate features enriched in successful HRL task-specific reasoning while filtering out source-language and generic-generation features.
We then construct steering directions from these features and inject them during LRL inference.
The resulting interventions test whether the selected features are functionally involved in the observed reasoning gap: suppressing them should impair source-language reasoning, while activating them should partially recover target-language reasoning beyond random and non-task controls.
Our framework reframes some cross-lingual reasoning gaps as failures of mechanism elicitation rather than capability absence, and offers a causally testable route to feature-mediated transfer without translation, fine-tuning, or changing the user-facing language.
\end{abstract}

%% file: sections/1_introduction.tex
\input{figures/motivation.tex}

\section{Introduction}
\label{sec:introduction}

Large language models (LLMs) have become increasingly capable multilingual systems~\cite{xue-etal-2021-mt5, workshop2023bloom176bparameteropenaccessmultilingual}, yet their reasoning abilities remain uneven across languages~\cite{bang-etal-2023-multitask, zhu-etal-2024-multilingual, kang2026multilingualreasoninggapsemerge}. 
For instance, a model that reliably solves a mathematical problem in a high-resource language (HRL), such as English or Spanish, may fail on an equivalent problem expressed in Thai, Bengali, Swahili, or another low-resource language (LRL).
This gap is often attributed to data imbalance: HRLs are better represented during pretraining, instruction tuning, and evaluation, leading to more robust behavior in those languages~\cite{joshi-etal-2020-state, nguyen2023culturaxcleanedenormousmultilingual}. For example, English alone makes up most of the pretraining data while low-resource languages get only a tiny share, and this gap shows up in their reasoning accuracy too (Figure~\ref{fig:motivation_example}).
While important, this explanation leaves a deeper question unresolved: 
\textit{Does language merely change the surface form of an input, or can it change which internal reasoning mechanisms a model uses?}

We study the hypothesis that language can modulate the elicitation of latent computation. 
Under this view, semantically equivalent prompts in different languages need not merely differ in surface form; they may induce different patterns of internal feature activation. 
An HRL may more reliably elicit computations useful for a task, such as symbolic decomposition or stepwise arithmetic reasoning, whereas an LRL may express the same problem while eliciting those computations more weakly, at different positions, or not at all~\cite{wendler-etal-2024-llamas,etxaniz-etal-2024-multilingual}. 
If this hypothesis is correct, cross-ingual performance gaps reflect not only differences in input distribution, but also differences in the accessibility of task-associated internal mechanisms.

We frame this problem as one of mechanistic intervention~\cite{elhage2021mathematical, meng2022locating, tang-etal-2024-language, li2023inferencetime}. 
Rather than asking only whether a model performs better in one language than another, we ask whether task-relevant latent features activated during successful HRL reasoning can be identified, tested, and reused during LRL inference. 
This perspective turns cross-lingual performance gaps into a causal question: \textit{which internal computations are activated, suppressed, or transferable across languages?}

To answer this question, we propose a sparse-feature intervention framework for cross-lingual reasoning. 
Using sparse autoencoders (SAEs)~\cite{ng2011sparse,chalnev2024improvingsteeringvectorstargeting,templeton2024scaling}, we represent model activations in an interpretable latent basis and construct task-specific feature libraries from successful HRL reasoning traces. 
Crucially, feature discovery is contrastive: selected features must be enriched during successful task execution while being less characteristic of generic HRL behavior. 
This helps distinguish reusable task mechanisms from features that merely encode language identity or broad generation patterns.

At inference time, we use the selected sparse features to construct a residual-stream steering direction from activation differences between HRL and LRL reasoning traces. 
The resulting vector is injected during LRL inference, leaving the prompt prefill unchanged while amplifying computations that are more strongly expressed in successful HRL reasoning. 
This allows us to test whether features associated with HRL success can causally improve LRL inference.

Our evaluation separates observational feature discovery from causal validation~\cite{10.5555/2074022.2074073}.
The contrastive procedure identifies candidate feature sets associated with successful reference-language (HRL) reasoning, but this association alone does not establish causality. 
We therefore evaluate the selected features through interventions in a fixed trained model: if the features are functionally involved in the relevant computation, activating them should improve target-language (LRL) inference, suppressing them should impair reference-language inference, and matched random or excluded-feature controls should not produce the same pattern.
This design does not estimate a formal natural indirect effect, but it provides interventional evidence that the selected features participate in a mechanism consistent with partial cross-lingual transfer. 
The resulting method requires no parameter updates, no translation at inference time, and no change to the user-facing language, though feature discovery uses benchmark correctness signals.

%% file: figures/motivation.tex
\begin{figure}[t]
\centering
\includegraphics[width=\columnwidth]{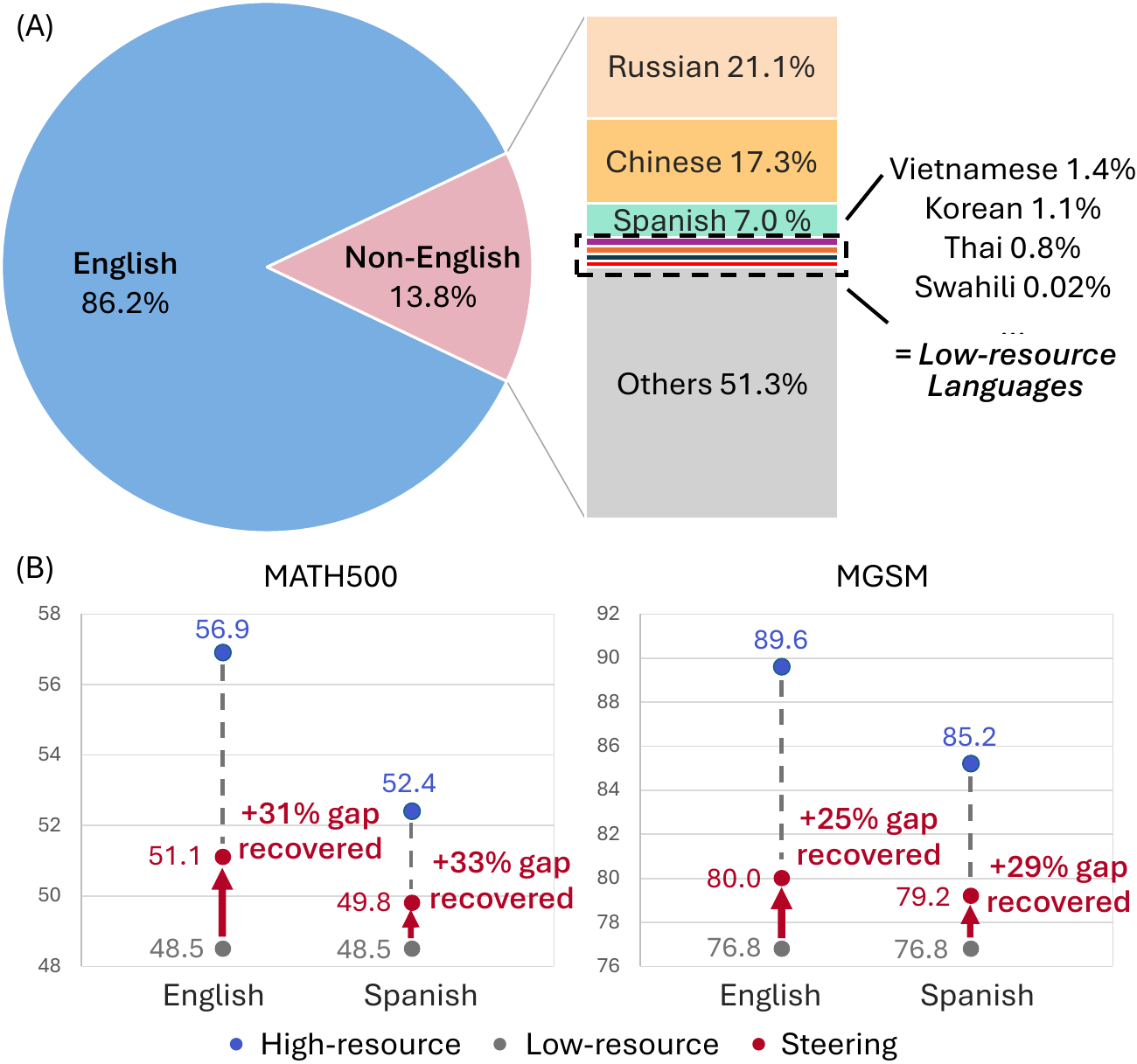}
\caption{ (A) In pretraining corpora FineWeb~\cite{penedo2024the},  English dominates the pretraining corpus (86.2\%), while target low-resource languages (Vietnamese, Korean, Thai, Swahili) together account for under 4\%. (B) On MATH500 and MGSM, Gemma-2-9B-it scores substantially higher when prompted in a high-resource reference language (blue: English or Spanish) than in low-resource targets (grey: averaged over Thai/Korean/Vietnamese for MATH500, and Thai/Korean/Swahili for MGSM). Steering with contrastively selected HRL reasoning features (red) recovers 25–33\% of the gap across all four (reference, benchmark) settings, without training or translation.}
\label{fig:motivation_example}
\vspace{-0.3cm}
\end{figure}

%% file: sections/2_related.tex
\section{Preliminaries}
\label{sec:preliminaries}

\subsection{Sparse Autoencoders (SAEs)}

SAEs provide a sparse decomposition of transformer activations into latent features. Given a residual-stream activation $h \in \mathbb{R}^{d}$ from a decoder-only language model, an SAE with $F$ latent features encodes $h$ into a sparse latent representation
\begin{equation}
    z = \mathrm{Enc}(h) = \phi\left(W_{\mathrm{enc}}(h - b_{\mathrm{dec}}) + b_{\mathrm{enc}}\right) \in \mathbb{R}^{F},
\end{equation}
where $W_{\mathrm{enc}} \in \mathbb{R}^{F \times d}$, $b_{\mathrm{enc}} \in \mathbb{R}^{F}$, $b_{\mathrm{dec}} \in \mathbb{R}^{d}$, and $\phi(\cdot)$ is a sparsity-inducing nonlinearity. The latent code is reconstructed through a linear decoder,
\begin{equation}
    \hat{h} = \mathrm{Dec}(z) = \sum_{f=1}^{F} z_f w_{\mathrm{dec}}^{(f)} + b_{\mathrm{dec}},
\end{equation}
where $w_{\mathrm{dec}}^{(f)} \in \mathbb{R}^{d}$ denotes the $f$-th column of the decoder matrix $W_{\mathrm{dec}} \in \mathbb{R}^{d \times F}$. Each decoder column corresponds to the residual-stream direction associated with latent feature $f$, and the coefficient $z_f$ determines the contribution of that feature to the reconstruction.

\subsection{Problem Setup and Notation}
Let $M$ be a decoder-only language model and let $\ell$ denote a residual-stream layer of interest. We attach a pretrained SAE to layer $\ell$ and analyze its latent representations during chain-of-thought (CoT) reasoning. We consider a reference-language (HRL) $b$ and a target-language (LRL) $a$.

Given a reasoning benchmark $\mathcal{D}$, the model produces a reasoning trace $r_i^{(k)}$ of length $T_i^{(k)}$ for problem $i \in \mathcal{D}$ in language $k \in \{a, b\}$. For each generation token position $t \in \{1, \dots, T_i^{(k)}\}$, let $h_{i,t}^{(k)} \in \mathbb{R}^d$ denote the residual-stream activation at layer $\ell$ and $z_{i,t}^{(k)} = \mathrm{Enc}(h_{i,t}^{(k)}) \in \mathbb{R}^{F}$ its SAE representation, with $z_{i,t,f}^{(k)}$ being the activation of latent feature $f$ at that token.

Our goal is to identify latent features associated with successful reasoning in the reference language $b$ and use them to steer the model toward similar behavior when solving problems in the target language $a$.

%%%%%%%%%%%%%%%%%

\subsection{Interventional Causal Effects} % inside a fixed model.
Our causal claims are restricted to interventions within a fixed trained model. 
Feature selection from naturally generated traces is observational: it identifies candidate internal variables associated with successful reasoning, but does not by itself establish causality. 

We therefore evaluate causal involvement by directly modifying residual-stream activations during the forward pass. 
For an internal activation $h_{\ell,t}$ and an intervention vector $\Delta h$, we write $Y\!\left(do(h_{\ell,t}\leftarrow h_{\ell,t}+\Delta h)\right)$ for the model's outcome under the intervention, where $Y$ denotes final-answer correctness or a logit-based answer score. 

A feature set is treated as \textit{sufficient} if activating its associated residual direction improves target-language (LRL) reasoning, \textit{necessary} if suppressing it degrades reference-language (HRL) reasoning, and \textit{specific} if matched random or excluded feature interventions do not produce the same effect.
We use the terms partial sufficiency and functional necessity operationally. 
A positive activation effect suggests that the corresponding residual directions are partially sufficient under the specified intervention policy; a negative suppression effect suggests functional dependence on the evaluated distribution. 
These terms do not imply logical sufficiency or necessity in the counterfactual-cause sense.

%% file: sections/3_method.tex
\input{figures/main}

\section{Method}
\label{sec:method}

This section details our method in two parts. First, we describe how we identify task-relevant sparse latent features through a contrastive comparison between reasoning traces of a high-resource reference language and a low-resource target language (LRL). Second, using those features, we intervene on the low-resource forward pass at a single residual stream layer to transfer reasoning behavior from the reference language.

\subsection{Task-Relevant Sparse Latent Feature Identification.}
\label{sec:feature_identification}

\paragraph{Paired contrast set and per-token features.}
Let $\mathcal{D}$ be a reasoning benchmark on which the model produces a chain-of-thought response $r_i^{(k)}$ for each problem $i$ and language $k$, scored as correct or incorrect. To isolate features that drive successful reasoning in a high-resource reference language $b$ but fail to engage in a low-resource target language (LRL) $a$, we restrict attention to the set of paired problems on which the reference succeeds and the target fails on the same question:
\begin{equation}
\mathcal{P}_{b \to a} = \bigl\{\, i \;\big|\; r_i^{(b)} \text{ correct} \,\wedge\, r_i^{(a)} \text{ incorrect} \,\bigr\}.
\label{eq:paired_set}
\end{equation}
For each $i \in \mathcal{P}_{b \to a}$ and each side $k \in \{a, b\}$, we record the identity of the strongest-activating feature at every generation token of the corresponding trace,
\begin{equation}
f^{*\,(k)}_{i,t} = \arg\max_{f \in \{1, \dots, F\}} z_{i,t,f}^{(k)},
\label{eq:top_feature}
\end{equation}
and define the per-response feature set $\mathcal{F}_i^{(k)} = \{f^{*\,(k)}_{i,t} : t = 1, \dots, T_i^{(k)}\}$ as the set of distinct top features across all generation tokens of trace $r_i^{(k)}$.

\paragraph{Set-difference candidate pool.}
For each feature $f$ and each side $k \in \{a, b\}$, we count the number of contrast responses in which it appears as the per-token argmax:
\begin{equation}
c_f^{(k)} = \bigl|\{\, i \in \mathcal{P}_{b \to a} : f \in \mathcal{F}_i^{(k)} \,\}\bigr|.
\label{eq:count}
\end{equation}
The contrastive candidate pool is then defined as the set-difference between the reference and target sides, retaining features that appear in the reference-correct responses but are absent from the matched target-incorrect ones:
\begin{equation}
\mathcal{C}_{\text{cand}} = \bigl\{\, f \;\big|\; c_f^{(b)} \geq 1 \,\wedge\, c_f^{(a)} = 0 \,\bigr\},
\label{eq:candidate_pool}
\end{equation}
ranked in descending order of $c_f^{(b)}$ so that higher ranks correspond to features that recur consistently across reference-correct traces.

\paragraph{Rank-window filtering.}
Features with the largest values in $\mathcal{C}_{\text{cand}}$ often correspond to highly frequent generation patterns that appear across reasoning traces regardless of their underlying reasoning content, providing limited signal for identifying reasoning-relevant latent dimensions. To reduce their influence, we retain only features whose ranks fall within a predefined window $(s,n)$, where $0 \le s < n$:
\begin{equation}
\mathcal{C}
=
\left\{
\mathcal{C}_{\text{cand}}^{(s+1)},
\mathcal{C}_{\text{cand}}^{(s+2)},
\dots,
\mathcal{C}_{\text{cand}}^{(n)}
\right\}.
\end{equation}

where $\mathcal{C}_{\text{cand}}^{(r)}$ denotes the feature at rank $r$ in the sorted candidate list.

\subsection{Transfer via Steering at the Residual Stream}

\paragraph{Per-feature steering coefficients.}
For each language $k \in \{a,b\}$, let $\mathcal{D}_k^+$ denote the set of correctly solved problems. We compute the mean SAE activation vector over all generated tokens from correctly solved reasoning traces:
\begin{equation}
\bar z^{(k)}
=
\frac{1}{N_k}
\sum_{i\in\mathcal D_k^+}
\sum_{t=1}^{T_i^{(k)}}
z_{i,t}^{(k)}
\in
\mathbb R^F,
\end{equation}
where $N_k = \sum_{i\in\mathcal D_k^+} T_i^{(k)}$ is the total number of generated tokens across all correctly solved traces in language $k$, and $z_{i,t}^{(k)}$ is the SAE representation defined in §\ref{sec:feature_identification}. For each selected feature $f\in\mathcal C$, we define its steering coefficient as the difference in mean activation between the reference language $b$ and the target language $a$:
\begin{equation}
w_f
=
\bar z_f^{(b)}
-
\bar z_f^{(a)}.
\end{equation}

Intuitively, $w_f$ measures how much more strongly feature $f$ is expressed in the reference language relative to the target language on successful reasoning traces.

\paragraph{Steering intervention.}
Let $h \in \mathbb{R}^d$ denote the layer-$\ell$ residual stream activation at a generation token, and let $z = \mathrm{Enc}(h) \in \mathbb{R}^F$ be its SAE representation. We construct a modified latent code by shifting the selected features according to their steering coefficients, where $\alpha>0$ controls the steering strength:

\begin{equation}
\tilde z_f=
\begin{cases}
z_f+\alpha w_f, & f\in\mathcal C,\\
z_f, & \text{otherwise},
\end{cases}
\end{equation}

The resulting intervention in residual space is given by
\begin{equation}
\Delta h
=
\mathrm{Dec}(\tilde z)
-
\mathrm{Dec}(z)
=
\alpha
\sum_{f\in\mathcal C}
w_f\,W_{\mathrm{dec}}[f],
\end{equation}
where the final equality follows from the linearity of the decoder. We register this intervention as a forward hook on the output of layer $\ell$ and apply it only to generation tokens, leaving the prompt prefill unchanged. The same steering vector $\Delta h$ is added at every generation step across all problems.

%% file: figures/main.tex
\begin{figure*}[t]
    \centering
    \includegraphics[width=\linewidth]{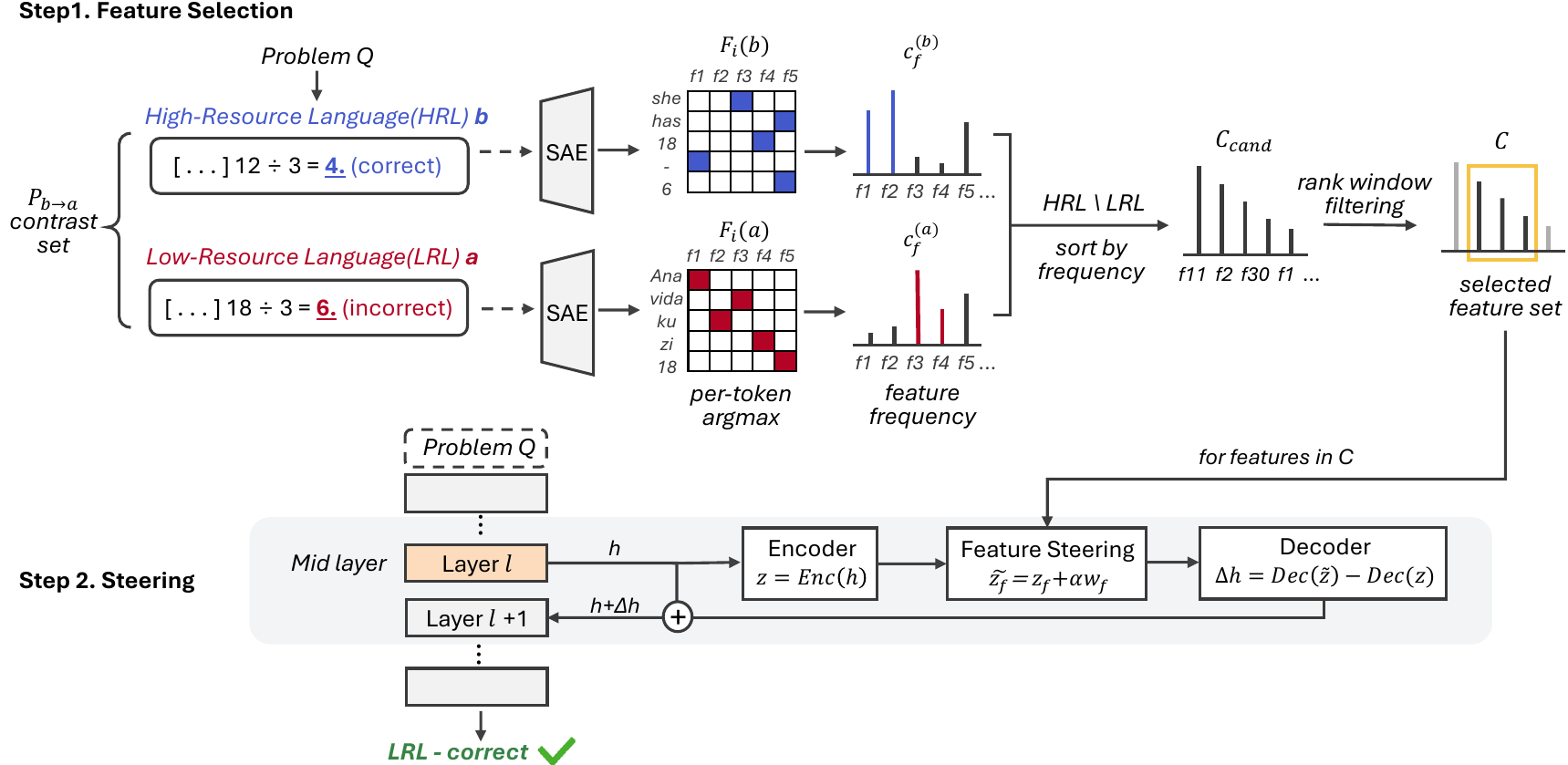}
    \caption{
        \textbf{Method overview.}
        \textbf{Step 1.} We collect the paired contrast set $\mathcal{P}_{b \to a}$ of problems on which the model is correct in the reference language $b$ but incorrect in the target language $a$. Encoding both traces through the SAE and counting the per-token argmax feature across the set gives per-side frequencies $c_f^{(b)}, c_f^{(a)}$; their set-difference followed by rank-window filtering yields the selected feature set $\mathcal{C}$.
        \textbf{Step 2.} At inference time on target-language inputs, the layer-$\ell$ residual stream $h$ is encoded to $z$, the features in $\mathcal{C}$ are bumped by $\alpha w_f$ where $w_f = \bar{z}_f^{(b)} - \bar{z}_f^{(a)}$, and the decoded difference $\Delta h$ is added back to $h$ at generation tokens only.
    }
    \label{fig:method_overview}
\end{figure*}

%% file: sections/4_experiments.tex
\section{Experiments}
\label{sec:experiments}

\input{tables/main_table}

\subsection{Experimental Setup}

We evaluate two instruction-tuned language models~\cite{gemmateam2024gemma2improvingopen,qwen2025qwen25technicalreport} paired with pretrained layer-20 Sparse Autoencoders (SAEs). For \texttt{gemma-2-9b-it}, we use the canonical Gemma-Scope SAE~\cite{lieberum2024gemmascopeopensparse} (\texttt{width\_16k}; $F=16{,}384$). For \texttt{Qwen2.5-7B-Instruct}, we use the released layer-20 matryoshka SAE~\footnote{\url{https://huggingface.co/chanind/qwen2.5-7B-it-layer-20-saes}} (\texttt{lmsys/matryoshka/k-100}; $F=65{,}536$, $k=100$). Both SAEs employ the JumpReLU architecture~\cite{rajamanoharan2024jumpingaheadimprovingreconstruction}. All steering interventions are applied at layer-20 residual stream.

\subsection{Datasets and Languages}

We evaluate on three multilingual reasoning benchmarks: \textbf{MATH500}~\cite{hendrycks2021measuring,lightman2023lets} ($N=311$), the MATH500 split of the multilingual MMATH benchmark~\citep{luo2025mmathmultilingualbenchmarkmathematical}, requiring multi-step symbolic reasoning; \textbf{MGSM}~\cite{shi2022languagemodelsmultilingualchainofthought} ($N=250$), a multilingual benchmark of grade-school mathematical word problems; and \textbf{MMLU-ProX (Psychology)}~\cite{xuan2025mmluproxmultilingualbenchmarkadvanced} ($N=798$), a 10-way multiple-choice benchmark covering undergraduate-level psychology questions.

Each problem is presented in a single language using a native-language chain-of-thought~\cite{wei2022chain} prompt. We consider English (en) and Spanish (es) as high-resource reference languages (HRL), and Korean (ko), Thai (th), Swahili (sw), and Vietnamese (vi) as target languages (LRL). Depending on the benchmark, translations are obtained from publicly available multilingual benchmark releases.

\subsection{Feature Selection and Steering}

For each reference--target language pair and dataset, we construct a contrastive feature pool using the procedure described in Section~\ref{sec:feature_identification}. Candidate features are identified from problems that are solved correctly in the reference language but incorrectly in the target language. Following rank-window filtering, we discard the top $50\%$ of candidate features ranked by occurrence count and retain features between the 50th and 90th percentiles. Only features that survive this filtering stage are used for steering.

During inference, we apply residual-stream steering at every generation step while leaving prompt-prefill activations unchanged.
\subsection{Evaluation Metrics}

\paragraph{Accuracy.}
Accuracy is computed using the native evaluation protocol of each benchmark. For MATH500 and MGSM, answers are evaluated using \texttt{math\_verify}; for MMLU-ProX, predictions are scored by matching the final selected answer option.

\paragraph{Recovery Rate.}
To quantify how much steering closes the performance gap between a target language and a reference language, we define the recovery rate

\begin{equation}
\mathrm{Recovery}_{a\rightarrow b}
=
\frac{
\mathrm{acc}_{a\rightarrow b}
-
\mathrm{acc}_{a}
}{
\mathrm{acc}_{b}
-
\mathrm{acc}_{a}
},
\end{equation}

where $\mathrm{acc}_{a}$ denotes the baseline accuracy of target language $a$, $\mathrm{acc}_{b}$ denotes the accuracy of reference language $b$, and $\mathrm{acc}_{a\rightarrow b}$ denotes the steered target-language accuracy.

%% file: tables/main_table.tex
% Requires: \usepackage{booktabs} \usepackage{multirow} \usepackage{graphicx}
\begin{table*}[t]
\centering
\resizebox{\textwidth}{!}{%
\begin{tabular}{ll cccc cccc ccccc}
\toprule
\multicolumn{15}{l}{\emph{Reference language: English}} \\
\midrule
\multirow{2}{*}{Model} & & \multicolumn{4}{c}{MATH500} & \multicolumn{4}{c}{MGSM} & \multicolumn{5}{c}{MMLU-ProX (psychology)} \\
\cmidrule(lr){3-6}\cmidrule(lr){7-10}\cmidrule(lr){11-15}
 & & en & th & ko & vi & en & th & ko & sw & en & th & ko & sw & vi \\
\midrule
\multirow{3}{*}{Gemma-2-9B} & Baseline  & 56.91 & 48.23 & 49.20 & 48.23 & 89.60 & 78.40 & 76.40 & 75.60 & 57.27 & 32.71 & 17.67 & 8.27 & 40.73 \\
 & +Steering &  & 49.20 & 52.41 & 51.77 &  & 82.40 & 78.40 & 79.20 &  & 33.30 & 23.31 & 10.30 & 42.36 \\
 & Recovery  &  & +11\% & +42\% & +41\% &  & +36\% & +15\% & +26\% &  & +2\% & +14\% & +4\% & +10\% \\
\midrule
\multirow{3}{*}{Qwen2.5-7B} & Baseline  & 73.63 & 60.77 & 64.95 & 63.67 & 95.20 & 80.40 & 79.60 & 15.20 & 59.90 & 39.35 & 42.48 & 15.29 & 49.50 \\
 & +Steering &  & 63.70 & 66.60 & 66.56 &  & 82.00 & 80.00 & 19.60 &  & 41.60 & 44.11 & 17.42 & 53.13 \\
 & Recovery  &  & +23\% & +19\% & +29\% &  & +11\% & +3\% & +6\% &  & +11\% & +9\% & +5\% & +35\% \\
\midrule\midrule
\multicolumn{15}{l}{\emph{Reference language: Spanish}} \\
\midrule
\multirow{2}{*}{Model} & & \multicolumn{4}{c}{MATH500} & \multicolumn{4}{c}{MGSM} & \multicolumn{5}{c}{MMLU-ProX (psychology)} \\
\cmidrule(lr){3-6}\cmidrule(lr){7-10}\cmidrule(lr){11-15}
 & & es & th & ko & vi & es & th & ko & sw & es & th & ko & sw & vi \\
\midrule
\multirow{3}{*}{Gemma-2-9B} & Baseline  & 52.41 & 48.23 & 49.20 & 48.23 & 85.20 & 78.40 & 76.40 & 75.60 & 45.49 & 32.71 & 17.67 & 8.27 & 40.73 \\
 & +Steering &  & 48.87 & 50.16 & 50.48 &  & 82.40 & 78.00 & 77.20 &  & 33.50 & 19.70 & 10.00 & 41.35 \\
 & Recovery  &  & +15\% & +30\% & +54\% &  & +59\% & +18\% & +17\% &  & +6\% & +7\% & +5\% & +13\% \\
\midrule
\multirow{3}{*}{Qwen2.5-7B} & Baseline  & 67.85 & 60.77 & 64.95 & 63.67 & 84.00 & 80.40 & 79.60 & 15.20 & 53.01 & 39.35 & 42.48 & 15.29 & 49.50 \\
 & +Steering &  & 63.34 & 65.90 & 66.56 &  & 82.00 & 82.00 & 19.60 &  & 41.10 & 43.61 & 15.54 & 51.38 \\
 & Recovery  &  & +36\% & +33\% & +69\% &  & +44\% & +55\% & +6\% &  & +13\% & +11\% & +1\% & +54\% \\
\bottomrule
\end{tabular}%
}
\caption{
\textbf{Main results: multilingual reasoning accuracy and cross-lingual gap closure.}
Accuracy (\%) on MATH500, MGSM, and MMLU-ProX (psychology subset) for Gemma-2-9B-it and Qwen2.5-7B-Instruct, with English (top) and Spanish (bottom) as reference languages.
For each (reference, target) pair, \textbf{Baseline} reports vanilla CoT accuracy, \textbf{+Steering} reports accuracy after applying our residual-stream intervention at layer 20, and \textbf{Recovery} is the fraction of the reference gap closed,
Reference-language scores (en, es columns of the Baseline row) serve as ceilings and are not steered.
Target language coverage varies by benchmark availability.
}
\label{tab:main_table}
\end{table*}

%% file: sections/5_discussion.tex
\section{Discussion}
\label{sec:discussion}

\input{tables/analysis_table}
\subsection{Causal Evidence for Feature-Mediated Transfer}

The contrastive feature selection procedure in Section~\ref{sec:method} identifies sparse features that are more salient in successful reference-language traces than in matched target-language failures. 
We treat the selected set $\mathcal{C}$ as a set of candidate intervention features: task-associated features that may transfer task-relevant computations from the reference language reasoning to the target language. 
We evaluate this interpretation with three intervention tests.

% \paragraph{Partial sufficiency.}
% For a target language $a$, we test whether activating a feature set $\mathcal{S}$ improves target-language reasoning:
% \begin{equation}
% \tau^a_S(\alpha)  = \mathbb{E}_i \left[ Y^a_i\!\left( \operatorname{do}(h_{\ell,t}\leftarrow h_{\ell,t}+\Delta h^+_S(\alpha)) \right) - Y^a_i \right],
% \end{equation}
% where $Y^a_i$ denotes final-answer correctness or a logit-based answer score.
% A positive effect for $\mathcal{S}=\mathcal{C}$ indicates that the selected feature directions are partially sufficient to recover target-language reasoning behavior under the intervention.
\paragraph{Partial sufficiency.}
For a target language $a$, we test whether activating a feature set $\mathcal{S}$ improves target-language reasoning:
\begin{align}
&\tau^a_S(\alpha) \\
&= \mathbb{E}_i \left[ Y^a_i\!\left( \operatorname{do}(h_{\ell,t}\leftarrow h_{\ell,t}+\Delta h^+_S(\alpha)) \right) - Y^a_i \right], \nonumber
\end{align}
where $Y^a_i$ denotes final-answer correctness or a logit-based answer score.
% A positive effect for $\mathcal{S}=\mathcal{C}$ indicates that the selected feature directions are partially sufficient to recover target-language reasoning behavior under the intervention.
When $\mathcal{S}$ is instantiated as the contrastively selected feature set $\mathcal{C}$, a positive value of $\tau^a_C(\alpha)$ indicates that activating these features improves target-language reasoning.
Table~\ref{tab:main_table} provides this target-side test: activating the pair-specific feature set $\mathcal{C}_{\mathrm{EN}\rightarrow a}$ improves accuracy over the corresponding LRL baseline for Thai, Korean, and Vietnamese.

These gains are largest for Korean and Vietnamese, indicating that the selected directions can partially recover reasoning behavior that is more reliably elicited in the English reference language.

Importantly, these gains do not generally arise from switching the generated output to the reference language. Automatic language identification with GlotLID~\citep{glotlid} shows that steering largely preserves target-language generation; detailed results across models, datasets, and reference--target pairs are reported in Appendix~\ref{app:language} (Table~\ref{tab:output_language}).

% \paragraph{Functional necessity.}
% For a reference language $b$, we test whether suppressing the same features degrades reference-language reasoning:
% \begin{align}
%     &\nu^b_S(\lambda)\\ &= \mathbb{E}_i \left[ Y^b_i\!\left(
% \operatorname{do}(h_{\ell,t}\leftarrow h_{\ell,t}+\Delta h^-_{S,t}(\lambda)) \right) - Y^b_i \right],  \nonumber 
% \end{align}
% where $\Delta h^-_{S,t}(\lambda))$ indicates anti-steering perturbation effect.
% If $\nu^b_C(\lambda)<0$, while matched random-feature suppression has a smaller effect, then the selected features are functionally involved in producing the reference-language behavior.
\paragraph{Functional necessity.}
For a reference language $b$, we test whether suppressing the same features degrades reference-language reasoning:
\begin{align}
    &\nu^b_S(\lambda)\\ &= \mathbb{E}_i \left[ Y^b_i\!\left(
\operatorname{do}(h_{\ell,t}\leftarrow h_{\ell,t}+\Delta h^-_{S,t}(\lambda)) \right) - Y^b_i \right],  \nonumber 
\end{align}
where $\Delta h^-_{S,t}(\lambda)$ indicates an anti-steering perturbation effect.
If $\nu^b_C(\lambda)<0$, while matched random-feature suppression has a smaller effect, then the selected features are functionally involved in producing the reference-language behavior.
The HRL ablation row in Table~\ref{tab:analysis_table} instantiates this test on the source side: for each target language $a$, we suppress the same pair-specific feature set $\mathcal{C}_{\mathrm{EN}\rightarrow a}$ during English inference.
Although the ablation is always performed in English, the Thai, Korean, and Vietnamese columns denote the target language used to construct $\mathcal{C}_{\mathrm{EN}\rightarrow a}$; the resulting drops from the English baseline show that these features are functionally involved in source-language reasoning.

% \paragraph{Specificity.}
% We compare $\mathcal{C}$ against matched random features $\mathcal{R}$ and excluded features $\mathcal{E}$.
% These controls test whether the observed effects are due to the selected task-associated features rather than arbitrary residual perturbation, intervention magnitude, or generic SAE feature manipulation.
\paragraph{Specificity.}
We compare $\mathcal{C}$ against matched random features $\mathcal{R}$ and excluded features $\mathcal{E}$.
These controls test whether the observed effects are due to the selected task-associated features rather than arbitrary residual perturbation, intervention magnitude, or generic SAE feature manipulation.
Table~\ref{tab:analysis_table} shows that top-$k$ frequent features and random steering do not reproduce the same consistent improvement pattern as the contrastively selected feature set.
Moreover, applying the negative steering vector reduces target-language accuracy, suggesting that the signed HRL--LRL activation difference is functionally meaningful rather than merely increasing residual-stream activation magnitude.

Together, the three tests provide interventional evidence that $\mathcal{C}$ is functionally implicated in the cross-lingual reasoning gap: activating $\mathcal{C}$ improves target-language performance, suppressing $\mathcal{C}$ degrades reference-language performance, and matched controls fail to reproduce either effect. 
Table~\ref{tab:analysis_table} directly operationalizes these criteria by evaluating the same pair-specific feature sets under target-side activation, source-side ablation, and control interventions.
We therefore interpret the selected sparse features as a mechanistic handle through which target-language inference can access computations more reliably elicited in the reference language.

\subsection{Are the steered features language-agnostic reasoning features?}
\label{sec:crosslingual}

\input{tables/feature_token}
To understand why injecting HRL-derived feature directions helps
LRL reasoning, we ask what the steered features actually encode. We take
the $(\mathrm{en},\mathrm{ko})$ contrastive pool used for steering and, for each feature, examine which tokens it activates on
in English versus Korean.

\paragraph{Setup.}
For 40 MGSM problems answered correctly in English, we perform a single forward pass on both the English and Korean reasoning traces, extract the layer-20 residual-stream activations, and encode them using the Gemma-Scope SAE. For each of the 28 selected features, we then identify the top-activating tokens and record the peak activation in each language.

\paragraph{The features are largely language-agnostic.}
Only $3$ of the $28$ features are English-token specific and do not activate on Korean text (\emph{of}, \emph{the}, \emph{to}); the remaining $25$ activate in both languages. 
% Crucially, a semantically interpretable subset corresponds to reasoning concepts whose surface form differs across languages yet which activate the same SAE latent in both (Table~\ref{tab:feature_token}): e.g.\
% feature~13089 fires on English \emph{needs/needed} and Korean \emph{필요}
% (``need''), feature~12672 on \emph{number/how~many} and \emph{수/몇}, and
% feature~4301 on \emph{equals/equation} and the symbol~\emph{=}. The remaining
% transferring features map onto shared syntactic categories (copulas, case
% particles, prepositions) rather than reasoning content.
Among these, a semantically interpretable subset captures reasoning concepts shared across languages while activating the same SAE latent (Table~\ref{tab:feature_token}). 

For example, feature~13089 activates on English \emph{needs/needed} and Korean \emph{필요} (``need''), feature~12672 on \emph{number/how~many} and \emph{수/몇}, and feature~4301 on \emph{equals/equation} and the symbol~\emph{=}. Other shared features correspond to syntactic categories, such as copulas, case particles, and prepositions, rather than task-specific reasoning content.

\paragraph{Interpretation.}

These observations help explain the steering result. If the features primarily captured English surface patterns, activating them during Korean inference would be unlikely to improve reasoning. Instead, they encode language-agnostic reasoning concepts shared across languages, such as quantity, requirement, equality, and aggregation. Moreover, these features typically activate \emph{more weakly} on Korean tokens than on their English counterparts (e.g., feature~13089 peaks at $28.0$ in Korean vs.\ $63.6$ in English). This pattern is consistent with our hypothesis that the model under-elicits task-relevant reasoning features in the low-resource language, and that steering can partially compensate for this activation gap.

% This explains the steering result. The injected directions are not detectors of
% English surface tokens; were they so, adding them to a Korean forward pass would
% be meaningless. Instead they encode language-agnostic reasoning concepts
% (quantity, requirement, equality, aggregation, \ldots) present in both languages,
% so steering amplifies reasoning content the target language already shares with
% the reference rather than importing English-specific signal. Moreover, each
% feature typically activates \emph{more weakly} on the Korean token than on its
% English counterpart (e.g.\ feature~13089 peaks at $28.0$ in Korean vs.\ $63.6$ in
% English; Table~\ref{tab:feature_token}), which is consistent with the model
% \emph{under-engaging} reasoning features in the low-resource language, the very
% gap our intervention is designed to close.

%% file: tables/analysis_table.tex
\begin{table}[t]
\centering
\small
\setlength{\tabcolsep}{4pt}
\begin{tabular}{p{3.0cm}cccc}
\toprule
\textbf{Method} & \textbf{EN} & \textbf{TH} & \textbf{KO} & \textbf{VI} \\
\midrule

Baseline
& 56.91 & 48.23 & 49.20 & 48.23 \\

Ours
& -- & 49.20 & 52.41 & 51.77 \\

Top-$k$ feature steering
& -- & 45.98 & 45.66 & 47.91 \\

Random steering
& -- & 47.80 & 46.84 & 52.30 \\

Negative steering vector
& -- & 46.62 & 47.27 & 46.95 \\

\midrule

HRL ablation
& --
& 54.02
& 50.16
& 52.09
\\
HRL ablation (random)
& --
& 55.16
& 55.09
& 55.52
\\

\bottomrule
\end{tabular}
\caption{
\textbf{Control and ablation experiments.}
We evaluate the contrastively selected feature set $\mathcal{C}$ against alternative steering choices and against a source-side suppression test.
\textbf{Ours}: steering with $\mathcal{C}$ (target-language accuracy).
\textbf{Top-$k$ feature steering}: steering with the most frequent features in $\mathcal{C}_{\text{cand}}$, skipping rank-window filtering.
\textbf{Random steering}: steering with a size-matched random feature set.
\textbf{Negative steering vector}: applying $-\Delta h$ instead of $+\Delta h$.
\textbf{HRL ablation}: suppressing $\mathcal{C}_{\text{EN}\to a}$ during English inference (rows labeled by target language $a$, but the score is on English).
All target-language steering results use Gemma-2-9B-it on MATH500 with English as reference.
}
\label{tab:analysis_table}
\end{table}

%% file: tables/feature_token.tex
\begin{table}[t]
\centering
\footnotesize
\setlength{\tabcolsep}{2pt}      % 컬럼 간격 축소
\begin{tabular}{@{}rlll@{}}
\toprule
Feat. & Concept & English (peak) & Korean (peak) \\
\midrule
13089 & requirement   & needs, needed (63.6)    & 필요 (28.0) \\
12672 & quantity      & number, many (51.0)     & 수, 몇 (23.5) \\
6224  & aggregation   & both, three (68.5)      & 모두 (17.9) \\
629   & knowledge     & know, understand (58.2) & 알 (27.0) \\
6763  & rate          & per, / (69.7)           & / (21.8) \\
4404  & construction  & make, makes (46.3)      & 만들 (12.2) \\
4301  & equality      & equals, equation (53.8) & = (11.4) \\
\bottomrule
\end{tabular}
\caption{Cross-lingual reasoning features in the $(\mathrm{en},\mathrm{ko})$
steering pool (\texttt{gemma-2-9b-it}, layer~20). For each feature we give the
concept it encodes and its top-activating tokens in English and Korean reasoning
traces (peak activation in parentheses). 
% The same SAE latent fires on the concept's realization in both languages.
}
\label{tab:feature_token}
\end{table}

%% file: sections/6_related_work.tex
\section{Related Work}
\label{sec:related_work}

\paragraph{Multilingual reasoning and cross-lingual transfer.}
Large language models are increasingly multilingual, yet they reason far better
in HRLs than in low-resource ones
\citep{bang-etal-2023-multitask, zhu-etal-2024-multilingual,
kang2026multilingualreasoninggapsemerge}. This gap is commonly attributed to the
imbalance of pretraining and instruction-tuning data across languages
\citep{joshi-etal-2020-state, penedo2024the}, and prior
remedies often rely on additional multilingual training
\citep{zhang2025lingualifteffectivetwostageinstruction}. A complementary line of
work studies multilingualism mechanistically: multilingual transformers appear to
process non-English inputs through a latent English-centric representation
\citep{wendler-etal-2024-llamas}, reasoning frequently improves when inputs are
routed through English \citep{etxaniz-etal-2024-multilingual}, and specific
neurons govern language-specific behaviour \citep{tang-etal-2024-language}. These
findings suggest cross-lingual gaps reflect not only the input distribution but
also which internal mechanisms a model engages, motivating an intervention-based
approach that requires no additional training.

\paragraph{Mechanistic interpretability and SAEs.}
Mechanistic interpretability studies LLM internals by decomposing models into interpretable units~\citep{olah2020zoom}, often validating hypotheses
through interventions such as activation patching and causal mediation analysis
that localize behaviour to specific components
\citep{meng2022locating, wang2022interpretability,
syed2023attribution}. Sparse autoencoders \citep{ng2011sparse, bricken2023monosemanticity,
cunningham2023sparseautoencodershighlyinterpretable} decompose the residual stream
of transformer models into sparse and often human-understandable features,
supporting the hypothesis that the latent space of LLMs is composed of linear and
interpretable directions \citep{arora, elhage2022toy}. We use the open Gemma~Scope
SAEs \citep{lieberum2024gemmascopeopensparse} to obtain such features and, in this
interventionist tradition, validate our selected features causally by activating,
suppressing, and substituting matched random or excluded controls.

\paragraph{Activation and feature steering.}
% Activation steering controls model outputs by adding a vector to the residual
% stream, typically the difference between activations on contrasting positive and
% negative prompts \citep{turner2024activationadditionsteeringlanguage}.
% \citet{panickssery2024steeringllama2contrastive} scaled this to Llama-2,
% \citet{repe} apply it across a wider range of tasks,
% \citet{cao2024personalizedsteeringlargelanguage} learn steering vectors that
% optimally drive desired outputs, and
% \citet{lee2024programmingrefusalconditionalactivation} introduce conditional
% steering. More recent
% work steers at the level of individual SAE features rather than raw activation
% differences: \citet{templeton2024scaling} clamp interpretable features,
% \citet{marks2024feature} edit sparse feature circuits, \citet{durmus2024steering}
% evaluate feature steering, and \citet{zhao2024steeringknowledgeselectionbehaviours}
% combine SAEs with representation engineering. We build on this direction,
% selecting reasoning features contrastively from successful high-resource traces
% and injecting them as a residual-stream direction during low-resource inference.

Activation steering controls model outputs by adding a vector to the residual stream, typically derived from contrasting positive and negative prompts \citep{turner2024activationadditionsteeringlanguage}. Subsequent work has scaled the approach to larger models \citep{panickssery2024steeringllama2contrastive}, broadened the range of behaviors it can target \citep{repe}, learned steering vectors directly for desired outputs \citep{cao2024personalizedsteeringlargelanguage}, and introduced input-conditional variants \citep{lee2024programmingrefusalconditionalactivation}. A complementary line of work steers at the level of individual SAE features, taking advantage of their interpretability to clamp or edit specific concepts \citep{templeton2024scaling, marks2024feature, durmus2024steering, zhao2024steeringknowledgeselectionbehaviours}. Our method belongs to this latter family, but differs in how features are selected: rather than relying on a single labeled concept or hand-chosen latents, we identify reasoning-relevant features contrastively from paired reference-correct, target-incorrect traces, and use the resulting set to bridge a \emph{cross-lingual} performance gap that prior steering work has not addressed.

%% file: sections/7_conclusion.tex
\section{Conclusion}
\label{sec:conclusion}

We have studied cross-lingual reasoning gaps in large language models from a mechanistic perspective, asking whether language modulates the elicitation of latent computations rather than only whether models perform better in some languages than others. Using sparse autoencoders, we identify task-relevant features through contrastive comparison between successful reference-language reasoning traces and matched target-language failures, and inject them at inference time as a residual-stream steering direction. Across two models, three benchmarks, and four low-resource target languages, this intervention closes 25--33\% of the gap to the reference language with no parameter updates, no translation, and no change to the user-facing language. Three intervention tests---partial sufficiency, functional necessity, and specificity---provide interventional evidence that the selected features are functionally implicated in the observed gap. Together, these results suggest that part of the cross-lingual reasoning gap reflects \emph{under-elicitation} of computations the model already possesses, rather than their absence, and point to feature-level intervention as a complementary route to translation- and training-based remedies.

%% file: sections/X_appendix.tex
\appendix

\section{Implementation Details}
\label{app:impl}

\paragraph{Models.}
We use two open instruction-tuned models: \texttt{google/gemma-2-9b-it}
(Gemma-2-9B-it) and \texttt{Qwen/Qwen2.5-7B-Instruct}. Both are loaded with HuggingFace transformers~\cite{DBLP:journals/corr/abs-1910-03771}. All interventions are
applied to the layer-20 residual stream.

\paragraph{Sparse autoencoders.}
For Gemma-2-9B-it we use the publicly released Gemma~Scope residual-stream SAEs
\citep{lieberum2024gemmascopeopensparse}
(\texttt{gemma-scope-9b-pt-res-canonical}, layer~20, width $16$k, i.e.\
$d_{\text{sae}}{=}16{,}384$; JumpReLU), loaded through \texttt{sae\_lens}. For
Qwen2.5-7B-Instruct we use the layer-20 matryoshka SAE
(\texttt{chanind/qwen2.5-7B-it-layer-20-saes}, \texttt{lmsys/matryoshka/k-100};
JumpReLU, $d_{\text{sae}}{=}65{,}536$), loaded with a manual configuration shim to
bridge \texttt{sae\_lens}~\cite{bloom2024saetrainingcodebase} config-schema versions. SAE activations are computed in
\texttt{float32} on the captured residual stream.

\paragraph{Software.}
All experiments use Python~3.11, PyTorch~2.4.1~\cite{paszke2019pytorch} (CUDA~12.1),
\texttt{transformers}~4.51.0, \texttt{sae\_lens}~5.5.2. Mathematical answers are scored with the
\texttt{math\_verify} library; multiple-choice (MMLU-ProX) answers use a dedicated
letter parser that extracts the leading letter from \verb|\boxed{...}| with an
``answer is X'' fallback.

\paragraph{Hardware.}
Experiments run on 4 NVIDIA H100 GPUs (80\,GB), using
a single GPU per job.

\paragraph{Inference and decoding.}
We decode greedily (\texttt{do\_sample=False}) with a maximum of $1{,}024$ new
tokens. A response is marked truncated if generation reaches
the token limit without emitting an end-of-sequence token. Prompts follow a
per-language native-language template instructing the model to reason step by step
and to place its final answer in \verb|\boxed{...}|.

\section{Output Language Consistency}
\label{app:language}

To verify that steering does not cause the model to revert to the
high-resource reference language, we perform automatic language identification
on all generated responses using GlotLID~\citep{glotlid}.
We measure the proportion of output identified as the intended target
low-resource language at two granularities: \textit{per-segment}, where each
response is divided into sentence-level segments and classified individually,
and \textit{per-response}, where the entire generated response receives a
single language label.

Table~\ref{tab:output_language} reports the target-language proportion before
and after steering for every model, dataset, reference language, and target
language combination. Overall, steering leaves the output language largely
unchanged from the baseline: at the segment level, the target-language proportion changes by less
than 1.5 percentage points in 37 out of 40 settings (average absolute change:
0.8 points; median: 0.41 points). This suggests that the observed accuracy
improvements arise primarily from improved reasoning within the target
language, rather than from the model reverting to the higher-resource
reference language.

\section{Additional Layer Interventions}
\label{app:layer}

We use layer~20 as the primary intervention point, motivated by prior findings that reasoning-relevant representations emerge prominently in intermediate layers~\cite{li-etal-2024-understanding,pmlr-v267-skean25a}. To test whether our results are specific to this choice, we additionally apply steering at layers~4, 12, 34, and~38 of Gemma-2-9B on MGSM, using English as the reference language.

As shown in Table~\ref{tab:layer_intervention}, steering improves performance over the baseline across a broad range of intervention layers. Layer~20 achieves the strongest overall performance across target languages, consistent with prior findings that intermediate layers contain strong task-relevant representations.

\section{Comparison with Related Steering Method}
\label{app:caa}

Contrastive Activation Addition (CAA)~\cite{caa} and our SAE-based approach both intervene on the residual stream at inference time. CAA operates directly on dense hidden-state differences, whereas our method steers sparse features associated with reasoning-relevant concepts shared across languages.

SAE-based steering achieves higher accuracy than CAA in 11 of the 12 evaluation settings while providing feature-level interpretability through individually identifiable features (Table~\ref{tab:caa_comparison}).

\section{Generalization to Unseen Target Languages} \label{app:cross_language} 
We further evaluate whether our method generalizes to target languages beyond those used in the main experiments. We consider Bengali (BN) and Telugu (TE) on MGSM with Gemma-2-9B, using English as the reference language. Pair-specific steering improves accuracy from 75.6\% to 81.2\% for Bengali and from 74.4\% to 75.6\% for Telugu, while preserving target-language generation (Table~\ref{tab:additional_languages}). 

We also test whether features identified for one target language can be reused for unseen target languages. Specifically, we apply feature sets extracted from EN$\rightarrow$KO and EN$\rightarrow$TH directly to Bengali and Telugu inference. Both transferred feature sets improve over the baseline, although pair-specific feature extraction remains strongest (Table~\ref{tab:cross_language}). 

We interpret this as evidence that each target language may under-elicit a partially different subset of reasoning features. Consequently, features selected for one language may be less relevant or already sufficiently active in another, leading to only partial cross-language transfer.
These results suggest that the feature sets are transferable across target languages, while pair-specific extraction better captures each language's own under-activated features. 

% ========================================================= 
% Tables 
% =========================================================

\begin{table*}[t]
\centering
\caption{
Target-language consistency before and after steering, measured using GlotLID. Values denote the percentage of generated output identified as the intended target language. Each cell reports \textbf{Baseline $\rightarrow$ Steered}.
}
\label{tab:output_language}
\small
\setlength{\tabcolsep}{6pt}
\renewcommand{\arraystretch}{1.08}

\begin{tabular}{llllcc}
\toprule
\textbf{Model} &
\textbf{Dataset} &
\textbf{Reference} &
\textbf{Target} &
\textbf{per-segment (\%)} &
\textbf{per-response (\%)} \\
\midrule

\multirow{20}{*}{Gemma-2-9B}
& \multirow{6}{*}{MATH500}
& EN & KO & 98.3 $\rightarrow$ 98.2 & 99.4 $\rightarrow$ 99.4 \\
& & EN & TH & 97.6 $\rightarrow$ 97.9 & 99.4 $\rightarrow$ 99.4 \\
& & EN & VI & 99.0 $\rightarrow$ 99.2 & 99.7 $\rightarrow$ 100.0 \\
& & ES & KO & 98.3 $\rightarrow$ 98.8 & 99.4 $\rightarrow$ 99.4 \\
& & ES & TH & 97.6 $\rightarrow$ 93.1 & 99.4 $\rightarrow$ 95.8 \\
& & ES & VI & 99.0 $\rightarrow$ 89.7 & 99.7 $\rightarrow$ 96.8 \\
\cmidrule(lr){2-6}

& \multirow{6}{*}{MGSM}
& EN & KO & 100.0 $\rightarrow$ 100.0 & 100.0 $\rightarrow$ 100.0 \\
& & EN & TH & 100.0 $\rightarrow$ 100.0 & 100.0 $\rightarrow$ 100.0 \\
& & EN & SW & 97.6 $\rightarrow$ 97.4 & 99.2 $\rightarrow$ 99.6 \\
& & ES & KO & 100.0 $\rightarrow$ 100.0 & 100.0 $\rightarrow$ 100.0 \\
& & ES & TH & 100.0 $\rightarrow$ 100.0 & 100.0 $\rightarrow$ 100.0 \\
& & ES & SW & 97.6 $\rightarrow$ 96.9 & 99.2 $\rightarrow$ 99.2 \\
\cmidrule(lr){2-6}

& \multirow{8}{*}{\shortstack[l]{MMLU-ProX\\(psychology)}}
& EN & KO & 99.9 $\rightarrow$ 99.9 & 100.0 $\rightarrow$ 99.9 \\
& & EN & TH & 99.2 $\rightarrow$ 99.1 & 99.6 $\rightarrow$ 99.6 \\
& & EN & VI & 98.9 $\rightarrow$ 99.8 & 99.5 $\rightarrow$ 99.9 \\
& & EN & SW & 99.4 $\rightarrow$ 99.4 & 99.8 $\rightarrow$ 99.8 \\
& & ES & KO & 99.9 $\rightarrow$ 99.9 & 100.0 $\rightarrow$ 100.0 \\
& & ES & TH & 99.2 $\rightarrow$ 98.2 & 99.6 $\rightarrow$ 99.9 \\
& & ES & VI & 98.9 $\rightarrow$ 99.5 & 99.5 $\rightarrow$ 99.8 \\
& & ES & SW & 99.4 $\rightarrow$ 99.2 & 99.8 $\rightarrow$ 99.8 \\

\midrule

\multirow{20}{*}{Qwen2.5-7B}
& \multirow{6}{*}{MATH500}
& EN & KO & 95.0 $\rightarrow$ 96.0 & 95.8 $\rightarrow$ 97.4 \\
& & EN & TH & 95.2 $\rightarrow$ 95.8 & 95.8 $\rightarrow$ 95.5 \\
& & EN & VI & 98.5 $\rightarrow$ 98.1 & 99.0 $\rightarrow$ 99.0 \\
& & ES & KO & 95.0 $\rightarrow$ 93.9 & 95.8 $\rightarrow$ 94.5 \\
& & ES & TH & 95.2 $\rightarrow$ 95.2 & 95.8 $\rightarrow$ 95.5 \\
& & ES & VI & 98.5 $\rightarrow$ 99.0 & 99.0 $\rightarrow$ 99.4 \\
\cmidrule(lr){2-6}

& \multirow{6}{*}{MGSM}
& EN & KO & 99.9 $\rightarrow$ 98.5 & 100.0 $\rightarrow$ 98.4 \\
& & EN & TH & 72.0 $\rightarrow$ 71.2 & 63.6 $\rightarrow$ 63.6 \\
& & EN & SW & 97.3 $\rightarrow$ 98.3 & 99.6 $\rightarrow$ 99.6 \\
& & ES & KO & 99.9 $\rightarrow$ 99.7 & 100.0 $\rightarrow$ 99.6 \\
& & ES & TH & 72.0 $\rightarrow$ 71.9 & 63.6 $\rightarrow$ 63.6 \\
& & ES & SW & 97.3 $\rightarrow$ 99.0 & 99.6 $\rightarrow$ 100.0 \\
\cmidrule(lr){2-6}

& \multirow{8}{*}{\shortstack[l]{MMLU-ProX\\(psychology)}}
& EN & KO & 96.7 $\rightarrow$ 95.7 & 97.0 $\rightarrow$ 95.5 \\
& & EN & TH & 93.0 $\rightarrow$ 92.0 & 91.7 $\rightarrow$ 90.6 \\
& & EN & VI & 99.7 $\rightarrow$ 99.4 & 99.8 $\rightarrow$ 99.6 \\
& & EN & SW & 92.1 $\rightarrow$ 93.0 & 96.1 $\rightarrow$ 94.9 \\
& & ES & KO & 96.7 $\rightarrow$ 95.8 & 97.0 $\rightarrow$ 95.7 \\
& & ES & TH & 93.0 $\rightarrow$ 93.1 & 91.7 $\rightarrow$ 91.7 \\
& & ES & VI & 99.7 $\rightarrow$ 99.7 & 99.8 $\rightarrow$ 99.9 \\
& & ES & SW & 92.1 $\rightarrow$ 91.6 & 96.1 $\rightarrow$ 95.2 \\

\bottomrule
\end{tabular}
\end{table*}

\begin{table*}
\centering
\caption{Accuracy (\%) on MGSM when steering Gemma-2-9B at different intervention layers, using English as the reference language.}
\label{tab:layer_intervention}
\small
\setlength{\tabcolsep}{7pt}
\begin{tabular}{lccc}
\toprule
\textbf{Layer} & \textbf{KO} & \textbf{TH} & \textbf{SW} \\
\midrule
Baseline & 76.4 & 78.4 & 75.6 \\
Layer 4  & 78.0 & 81.6 & 76.8 \\
Layer 12 & \textbf{79.2} & 80.8 & 76.0 \\
Layer 20 & 78.0 & \textbf{82.4} & \textbf{79.2} \\
Layer 34 & 78.0 & 81.6 & 77.2 \\
Layer 38 & 75.2 & 81.2 & 76.8 \\
\bottomrule
\end{tabular}
\end{table*}

\begin{table*}[t]
\centering
\caption{Comparison with CAA on Gemma-2-9B. Values report accuracy (\%). Best steering results are shown in bold.}
\label{tab:caa_comparison}
\small
\setlength{\tabcolsep}{4.5pt}
\begin{tabular}{llcccccc}
\toprule
& & \multicolumn{3}{c}{\textbf{MATH500}} & \multicolumn{3}{c}{\textbf{MGSM}} \\
\cmidrule(lr){3-5} \cmidrule(lr){6-8}
\textbf{Ref.} & \textbf{Method} & \textbf{TH} & \textbf{KO} & \textbf{VI} & \textbf{TH} & \textbf{KO} & \textbf{SW} \\
\midrule
\multirow{3}{*}{EN}
& Baseline & 48.23 & 49.20 & 48.23 & 78.40 & 76.40 & 75.60 \\
& CAA      & 46.95 & 47.91 & 51.45 & 80.40 & 77.20 & \textbf{80.00} \\
& Ours     & \textbf{49.20} & \textbf{52.41} & \textbf{51.77} & \textbf{82.40} & \textbf{78.40} & 79.20 \\
\midrule
\multirow{3}{*}{ES}
& Baseline & 48.23 & 49.20 & 48.23 & 78.40 & 76.40 & 75.60 \\
& CAA      & 44.40 & 48.60 & 48.90 & 80.80 & 77.20 & 76.00 \\
& Ours     & \textbf{48.87} & \textbf{50.16} & \textbf{50.48} & \textbf{82.40} & \textbf{78.00} & \textbf{77.20} \\
\bottomrule
\end{tabular}
\end{table*}

% \clearpage

\begin{table*}[t]
\centering
\caption{Generalization to additional low-resource target languages on MGSM with Gemma-2-9B. Recovery denotes the fraction of the performance gap to the English reference recovered by steering.}
\label{tab:additional_languages}
\small
\setlength{\tabcolsep}{8pt}
\begin{tabular}{llcc}
\toprule
\textbf{Metric} & \textbf{Method} & \textbf{BN} & \textbf{TE} \\
\midrule
\multirow{3}{*}{Accuracy (\%)}
& Baseline  & 75.6 & 74.4 \\
& +Steering & \textbf{81.2} & \textbf{75.6} \\
& Recovery  & +40\% & +8\% \\
\midrule
\multirow{2}{*}{Target language proportion (\%)}
& Baseline  & 99.76 & 99.88 \\
& +Steering & 100.00 & 99.63 \\
\bottomrule
\end{tabular}
\end{table*}

\begin{table*}[t]
\centering
\caption{
Cross-language transfer to unseen target languages on MGSM with Gemma-2-9B. Values report accuracy (\%). Target-specific features are extracted separately for each target language, while EN$\rightarrow$KO and EN$\rightarrow$TH features are reused directly for BN and TE.
}
\label{tab:cross_language}
\small
\setlength{\tabcolsep}{8pt}
\begin{tabular}{lcc}
\toprule
\textbf{Feature Set} & \textbf{BN} & \textbf{TE} \\
\midrule
Baseline                  & 75.6 & 74.4 \\
Target-specific             & \textbf{81.2} & \textbf{75.6} \\
EN$\rightarrow$KO features & 78.8 & 74.8 \\
EN$\rightarrow$TH features & 78.8 & 75.2 \\
\bottomrule
\end{tabular}
\end{table*}